\documentclass[preprint,12pt]{elsarticle}

\usepackage{amsmath,amsfonts,bm}

\def\eqref#1{equation~\ref{#1}}

\def\1{\bm{1}}

\def\vh{{\bm{h}}}

\def\vv{{\bm{v}}}

\def\vx{{\bm{x}}}
\def\vy{{\bm{y}}}

\def\mA{{\bm{A}}}

\def\mD{{\bm{D}}}

\def\mH{{\bm{H}}}
\def\mI{{\bm{I}}}

\def\mW{{\bm{W}}}

\DeclareMathAlphabet{\mathsfit}{\encodingdefault}{\sfdefault}{m}{sl}
\SetMathAlphabet{\mathsfit}{bold}{\encodingdefault}{\sfdefault}{bx}{n}

\usepackage{lineno,hyperref}
\usepackage{booktabs}
\modulolinenumbers[5]
\usepackage{amsfonts}

\usepackage{paralist}
\usepackage{amsmath}
\usepackage{graphicx}
\usepackage{pdfpages}
\usepackage{subfigure}
\usepackage{xspace}
\usepackage{multirow}
\usepackage{color}
\usepackage{url}
\usepackage{flushend}
\usepackage[skip=0pt]{caption}
\usepackage{balance}
\usepackage{cleveref}
\usepackage{float}

\usepackage{array}

\usepackage{subfigure}

\usepackage{paralist}
\usepackage[numbers]{natbib}

\newcommand{\myfootnote}[1]{\footnote{\small #1}}
\newcommand{\myparagraph}[1]{\vspace{1ex}\noindent\textbf{#1.}\hspace{1em}}

\newcommand{\our}{\textsf{HMEA}\xspace}
\newcommand{\densenet}{\textsf{densenet}\xspace}
\newcommand{\hgcn}{\textsf{HGCNs}\xspace}
\newcommand{\gcn}{\textsf{GCN}\xspace}
\newcommand{\poe}{\textsf{PoE}\xspace}
\newcommand{\gcnalign}{\textsf{GCN-align}\xspace}
\newcommand{\dpg}{\textsf{DPGCNN}\xspace}
\newcommand{\gat}{\textsf{GAT}\xspace}
\newcommand{\trans}{\textsf{TransE}\xspace}
\newcommand{\gnn}{\textsf{GNN}\xspace}
\newcommand{\rdgcn}{\textsf{RDGCN}\xspace}
\newcommand{\mugnn}{\textsf{MuGNN}\xspace}
\newcommand{\kecg}{\textsf{KECG}\xspace}
\newcommand{\vgg}{\textsf{VGG16}\xspace}
\newcommand{\ikrl}{\textsf{IKRL}\xspace}

\newcommand{\dbpsz}{$\texttt{DBP15K}_\texttt{ZH-EN}$\xspace}
\newcommand{\dbpsj}{$\texttt{DBP15K}_\texttt{JA-EN}$\xspace}
\newcommand{\dbpsf}{$\texttt{DBP15K}_\texttt{FR-EN}$\xspace}

\newtheorem{Example}{\textbf{Example}}

\journal{Neurocomputing}

\begin{document}

\begin{frontmatter}



\thispagestyle{empty}
\title{Multi-modal Entity Alignment in Hyperbolic Space}


\author[]{Hao Guo}
\author[]{Jiuyang Tang}
\author[]{Weixin Zeng}
\author[]{Xiang Zhao}
\author[]{Li Liu}

\address{National University of Defense Technology, China}

\begin{abstract}

Many AI-related tasks involve the interactions of data in multiple modalities. It has been a new trend to merge multi-modal information into knowledge graph(KG), resulting in multi-modal knowledge graphs (MMKG). 
However, MMKGs usually suffer from low coverage and incompleteness. 
To mitigate this problem, a viable approach is to integrate complementary knowledge from other MMKGs. 
To this end, although existing entity alignment approaches could be adopted, they operate in the Euclidean space, and the resulting Euclidean entity representations can lead to large distortion of KG's hierarchical structure. Besides, the visual information has yet not been well exploited.

In response to these issues, in this work, we propose a novel multi-modal entity alignment approach, Hyperbolic multi-modal entity alignment (\our), which extends the Euclidean representation to hyperboloid manifold. 
We first adopt the Hyperbolic Graph Convolutional Networks (\hgcn) to learn structural representations of entities.
Regarding the visual information, we generate image embeddings using the \densenet model, which are also projected into the hyperbolic space using \hgcn. 
Finally, we combine the structure and visual representations in the hyperbolic space and use the aggregated embeddings to predict potential alignment results. 
Extensive experiments and ablation studies demonstrate the effectiveness of our proposed model and its components.

\end{abstract}


\end{frontmatter}


\section{Introduction}
Over recent years, knowledge graph (KG) has become a popular data structure for representing factual knowledge in the form of RDF triples, which can facilitate a pile of downstream applications such as question answering~\cite{cui2019kbqa}, information extraction~\cite{yao2014information}, etc. 
Currently, we have a large number of general KGs (e.g., DBpedia~\cite{auer2007dbpedia}, YAGO~\cite{suchanek2007yago}, Google's Knowledge Vault~\cite{dong2014knowledge}), and domain-specific KGs (e.g., Medical and Molecule KGs).

\begin{figure}[t]
	\centering
	\includegraphics[width=1.0\linewidth]{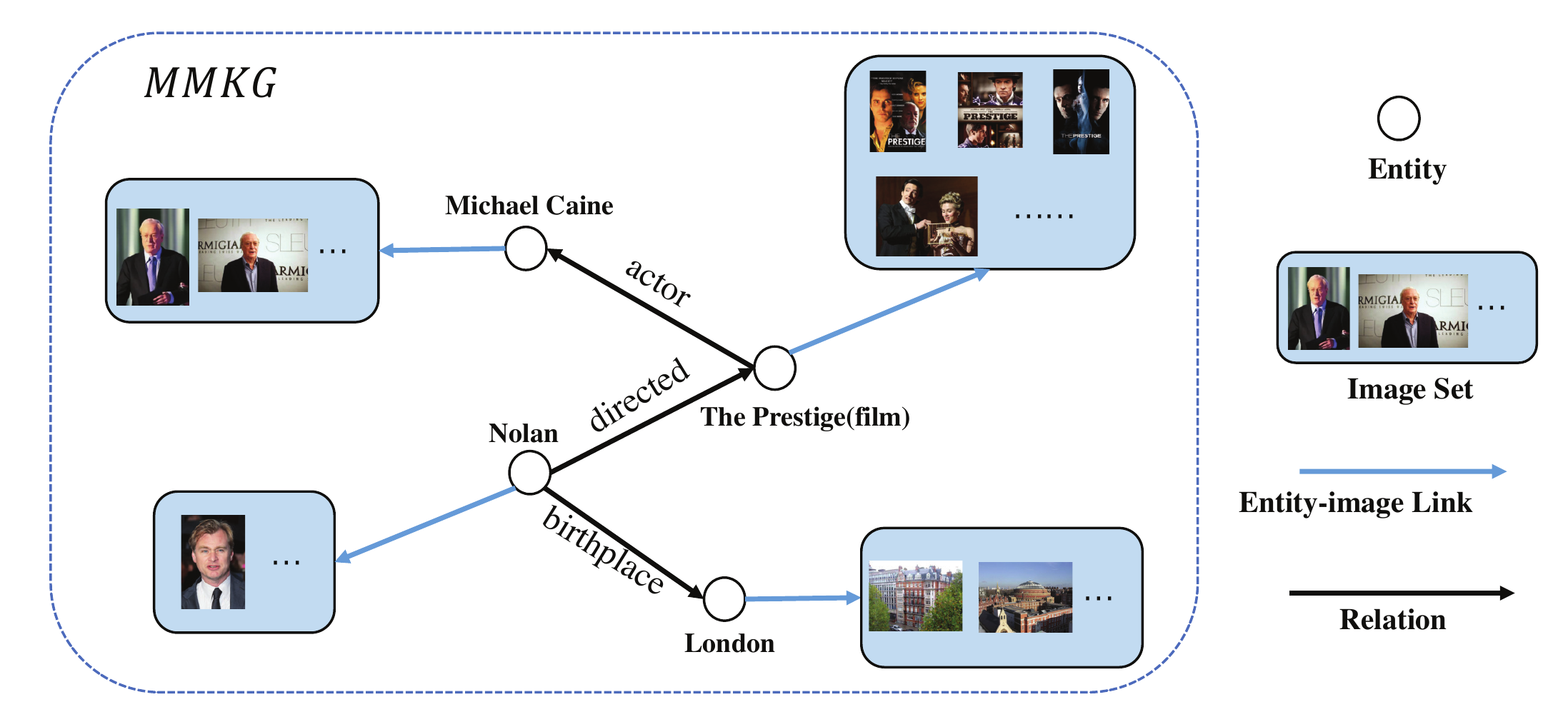}
	\caption{An example of MMKG.}
	\label{fig:mmkg}
\end{figure}

Meanwhile, there is a growing trend to incorporate multi-media information into KGs, so as to support cross-modal tasks that involve the interactions of data in multiple modalities, e.g., image and video retrieval~\cite{veltkamp2013state}, video summaries~\cite{newman2017generating}, visual entity disambiguation~\cite{moon2018multimodal}, and visual question answering~\cite{yi2018neural}, etc. 
To this end, several multi-modal KGs (MMKGs)~\cite{liu2019mmkg,wang2019richpedia} have been constructed very recently. 
An example of MMKG can be found in Figure~\ref{fig:mmkg}. 
In this work, without losing generality, we consider MMKG with two modalities, i.e., the KG structure information and visual information.

\begin{Example}
\label{egintto}
Figure~\ref{fig:mmkg} shows a partial MMKG, which consists of entities, image sets and the links between them. 
Specifically, the KG structure information includes the relations among entities, while the visual information comes from the image sets. For the entity $The Prestige$, its image set may contain scenes, actors, posters, etc. 
\end{Example}

Nevertheless, existing MMKGs usually come from limited data sources, and hence might suffer from low coverage of the domain~\cite{paulheim2017knowledge}. 
To improve the coverage of these MMKGs, a possible approach is to integrate useful knowledge from other MMKGs. 
Particularly, identifying equivalent entities in different KGs is a pivotal step to consolidate the knowledge among MMKGs, since entities are the anchors that connect these heterogeneous KGs. 
This process is also termed as multi-modal entity alignment (MMEA). 

MMEA is a non-trivial task, as it requires the modeling and integration of multi-modal information. 
For the \emph{KG structure information}, existing entity alignment (EA) approaches~\cite{hao2016joint,chen2016multilingual,DBLP:conf/ijcai/SunHZQ18,DBLP:conf/icde/Zeng0T020} can be directly adopted to generate entity structural embeddings for MMEA. 
These methods usually utilize \trans-based or graph convolutional network(\gcn)-based models~\cite{DBLP:conf/nips/BordesUGWY13,DBLP:journals/corr/KipfW16} to learn entity representations of individual KGs, which are then unified using the seed entity pairs. 
Nevertheless, all of these methods learn entity representations in the Euclidean space, which leads to a large distortion when embedding real-world graphs with scale-free or hierarchical structure~\cite{chen2013hyperbolicity, ravasz2003hierarchical}.
Regarding the \emph{visual information}, the \vgg model has been harnessed to learn the embeddings of images associated with entities and then used for alignment. However, the \vgg model cannot sufficiently extract useful features from images, which in turn constrains the effectiveness of alignment. 
Last but not least, the information from these two modalities should be carefully integrated so as to improve the overall effectiveness.


To address the aforementioned issues, in this work, we propose a multi-modal entity alignment approach that operates in Hyperbolic Space (\our). 
Specifically, we extend the Euclidean representation to hyperboloid manifold and adopt the Hyperbolic Graph Convolutional Networks (\hgcn) to learn structural representations of entities.
Regarding the visual information, we generate image embeddings using the \densenet model, which are also projected into the Hyperbolic space using \hgcn. 
Finally, we merge the structure embeddings and image embeddings in the hyperbolic space to predict potential alignments.


In summary, the major contributions of our approach can be summarized as follows:
\begin{itemize}
	\item We propose a novel MMEA approach, \our, which models and integrates multi-modal information in the hyperbolic space. 
	\item We adopt the Hyperbolic Graph Convolutional Networks (HGCNs) to learn structural representations of entities and demonstrate the advantage of Hyperbolic space for knowledge graph representations.
	\item We utilize a more advanced image embedding model to learn better visual representations for alignment. 	
	\item We validate the effectiveness of our proposed model via comprehensive experimental evaluations. 
\end{itemize}

\paragraph{Organization}
Section 2 overviews related work, and the preliminaries are introduced in Section 3. 
Section 4 describes our proposed approach. Section 5 presents experimental results, followed by conclusion in Section 6.

\section{Related Work}
In this section, we introduce some efforts that are relevant to this work.

\subsection{Multi-modal Knowledge Graph}

Most of the knowledge graph construction works focus on organizing and discovering textual knowledge in a structured representation, while paying little attention to other types of resources on the Web~\cite{wang2019richpedia}. Nevertheless, real-life applications involve cross-modal data, such as image and video retrieval, video summaries, visual question answering, visual commonsense reasoning, etc. To this end, Multi-modal Knowledge Graphs(MMKGs) are put forward, which contain various information (i.e. image, text, KG) and cross-modal relations. However, there are several challenges in building MMKGs. 
Extracting massive data of multiple modalities from search engines to build them is a time-consuming and labor-intensive project. In addition,  MMKGs usually suffer from low coverage of the domain and are incomplete. 
Integrating multi-modal knowledge from other MMKGs is an efficient approach to improve its completeness. 
Currently, there are few studies about merging different MMKGs. Liu et al.~\cite{liu2019mmkg} built two pairs of MMKGs, and extracted relational, latent, numerical and visual features for predicting the $SameAs$ link between entities. And some approaches of multi-modal knowledge representation involve visual features from entity images for knowledge repesentation learning, \ikrl~\cite{xie2017image-embodied} integrates image representations into an aggregated image-based representation via an attention-based method.

\subsection{Entity Alignment}
Over recent years, there are many works dedicated to the task of entity alignment~\cite{ji2020survey}, which could be used to model the structual information of MMKGs. 
Entity alignment is the task of finding equivalent entities in two KGs that refer to the same real-world object, which plays a pivotal step in automatically consolidating knowledge among KGs~\cite{chen2016multilingual,hao2016joint,DBLP:conf/sigir/Zeng00TT20}.
In general, current EA approaches mainly tackle the problem by assuming that equivalent entities in different KGs possess similar neighboring structure, and employing representation learning methods to embed entities as data points in a low-dimensional feature space. The distance between data points is used to evaluate the similarity of corresponding entities.

There are mainly two types of embedding learning models. \trans-based models~\cite{bordes2013translating} suggest that the embedding of the tail entity $t$ should be close to the embedding of the head entity $h$ plus the embedding of the relationship $r$ .
Meanwhile, Graph neural networks (\gnn) have received increased attention due to their attractive properties for learning from graph-structured data~\cite{bronstein2017geometric}.
Originally proposed in~\cite{gori2005new}, as a method for learning node representations on graphs using neural networks, this idea was extended to convolutional neural networks using spectral methods~\cite{defferrard2016convolutional}. Graph convolutional network(\gcn) can directly operate on graph-structured data and generate node-level embeddings by encoding the information about node neighborhoods.
\gcnalign~\cite{wang2018cross} uses \gcn to learn the structure and attribute information, then combines them with a balanced weight.
Noticing that \gcn neglects the relations in KGs, \rdgcn~\cite{wu2019relation} adopts the dual-primal graph convolutional neural network (\dpg)~\cite{monti2018dual} as a remedy. \mugnn~\cite{cao2019multi}, on the other hand, utilizes an attention-based \gnn model to assign different weights to different neighboring nodes. \kecg~\cite{li2019semi} combines the graph attention network (\gat)~\cite{vaswani2017attention} and \trans to capture both the inner-graph structure and the inter-graph alignment information.

\subsection{Representation Learning in Hyperbolic Space}

Essentially, most of the existing GCNs models are designed for the graphs in Euclidean spaces. ~\cite{cavallari2019embedding}
However, some works have discovered that graph data exhibits a non-Euclidean latent anatomy~\cite{muscoloni2017machine}, and embedding real-world graphs with scale-free or hierarchical structure leads to a large distortion~\cite{chen2013hyperbolicity, ravasz2003hierarchical}. Additionally, several recent researches in network science also show that hyperbolic geometry in particular is well-suited for modeling complex networks, as the hyperbolic space may reflect some properties of graph naturally~\cite{krioukov2010hyperbolic}. 
One key property of hyperbolic spaces is that they expand faster than Euclidean spaces, because Euclidean spaces expand polynomially while hyperbolic spaces expand exponentially.
Due to the natural advantage of hyperbolic space for graph structure data representation, representation learning in hyperbolic spaces has received increasing attention, especially on learning the hierarchical representation of a graph~\cite{nickel2017poincare}. 
Furthermore, Nickel et al.~\cite{nickel2018learning} showed that the Lorentz model of hyperbolic geometry has attractive properties for stochastic optimization and leads to substantially improved embeddings, especially in low dimensions.
Besides, some researches began to extend deep learning methods to hyperbolic space and showed state-of-the-art performance on link prediction and node classification tasks~\cite{ganea2018hyperbolic,gulcehre2018hyperbolic,tran2018survey}.

\section{Preliminaries}
In this section, we first formally define the task of MMEA, and then we briefly review the GCN model. Finally, we introduce the basic concepts of hyperbolic geometry, which serve as building blocks for our proposed model.

\subsection{Task Formulation}
MMEA aims to align entities in two MMKGs. A MMKG usually contains information in multiple modalities. In this work, without loss of generality, we focus on the KG structure information and visual information.
Formally, we represent a MMKGs as $MG = (E,R,T,I)$, where $E$, $R$, $T$, and $I$ denote the sets of entities, relations, triples and images, respectively. 
A relational triple $t \in T$ can be represented as $(e_1, r, e_2)$, where $e_1, e_2 \in E$ and $r \in R$. 
An entity $e$ is associated with multiple images $I_e = \{i_e^0, i_e^1,...,i_e^n\}$. 

\begin{figure}[t]
\centering
\includegraphics[width=\linewidth]{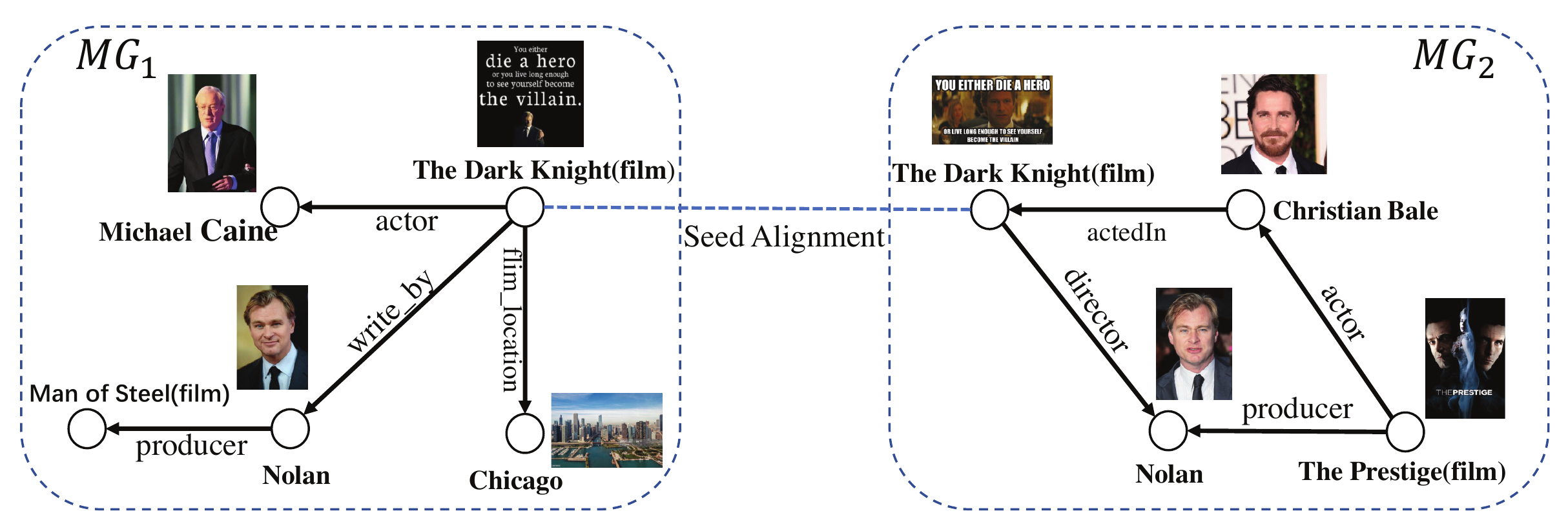}
\caption{An example of MMEA. Seed entity pairs are connected by dashed lines. For clarity,  we only choose an image to represent the set of images of an entity.}
\label{fig:1}
\end{figure}

Given two MMKGs, $MG_1 = (E_1, R_1, T_1, I_1)$, $MG_2 = (E_2, R_2, T_2, I_2)$, and seed entity pairs (pre-aligned entity pairs for training) $ S=\{(e_s^1, e_s^2)|e_s^1\leftrightarrow e_s^2, e_s^1 \in E_1,  e_s^2 \in E_2 \}$, where $\leftrightarrow$ represents equivalence, the task of MMEA can be defined as discovering more aligned entity pairs $\{(e^1, e^2)|e^1 \in E_1,  e^2 \in E_2 \}$. 
We use Example~\ref{eg} to further illustrate this task.

\begin{Example}
	\label{eg}
	Figure~\ref{fig:1} shows two partial MMKGs. The equivalence between \texttt{The Dark Knight} in $MG_1$ and \texttt{The Dark Knight} in $MG_2$ is known in advance. EA aims to detect potential equivalent entity pairs, e.g., \texttt{Nolan} in $MG_1$ and \texttt{Nolan} in $MG_2$, using the known alignments.
\end{Example}

\subsection{Graph Convolutional Neural Networks}
GCNs~\cite{henaff2015deep, kipf2016semi} are a type of neural network that directly operates on graph data. 
A GCN model consists of multiple stacked GCN layers. 
The inputs to the $l$-th layer of the GCN model are feature vectors of nodes and the structure of the graph.
$\mH^{(l)} \in {R} ^ {n \times d^{l}}$ is a vertex feature representation, where $n$ is the number of vertices and $d^{l}$ is the dimensionality of feature matrix. 
$\hat \mA = \mD^{-\frac{1}{2}}(\mA + \mI)\mD^{-\frac{1}{2}}$ represents the symmetric normalized adjacency matrix.
The identity matrix $\mI$ is added to the adjacency matrix $\mA$ to obtain self-loops for each node, and the degree matrix $ \mD = \sum _j(\mA_{ij}+\mI_{ij})$.
The output of the $l$-th layer is a new feature matrix $\mH^{(l+1)}$ by the following convolutional computation: 
\begin{equation}
 \mH^{(l+1)} = \sigma ( \hat \mA \mH^{(l)} \mW^{(l)}).
\end{equation}

\subsection{Hyperboloid Manifold}
We briefly review the key notions from the hyperbolic geometry; a more in-depth description is available in~\cite{eisenhart2015introduction}.
Hyperbolic geometry is a non-Euclidean geometry with constant negative curvature that measures how a geometric object deviates from a flat plane.
In this work, we use the $d$-dimensional Poincare ball model with negative curvature -$c$ $(c > 0)$:
$P^{(d,c)}=\{ \mathbf{x} \in R^d: \| \mathbf{x} \|^2 < \frac{1}{c}\}$, 
where $\| \cdot \|$ is the $L_2$ norm. 
For each point $ x \in P^{(d,c)}$, the tangent space $T^c_x $ is a $d$-dimensional vector space at point $x$, which contains all possible directions of paths in $P^{(d,c)}$ leaving from $x$.
Then, we introduce some basic operations in the hyperbolic space, which are essential in our proposed model. 

\myparagraph{Exponential and logarithmic maps}Specifically, let $\vv$ be the feature vector in the tangent space $ T^c_\mathbf{o}$; $\mathbf{o}$ is a point in the hyperbolic space $P^{(d,c)}$, which is also used as a reference point. Let $\mathbf{o}$ be the origin, $\mathbf{o} = 0$.
The tangent space $ T^c_\mathbf{o} $ can be mapped to $P^{(d,c)}$ via the exponential map:
\begin{equation}
 exp_\mathbf{o}^c(\vv) = \operatorname{tanh}(\sqrt{c}\|\vv\|)\frac{\vv}{\sqrt{c}\|\vv\|}
 \label{exp}.
\end{equation}

And conversely, the logarithmic map which maps  $P^{(d,c)}$ to  $ T^c_\mathbf{o} $ is defined as:
\begin{equation}
 log_\mathbf{o}^c({\vy}) = arctanh(\sqrt{c}\|\vy\|)\frac{{\vy}}{\sqrt{c}\|\vy\|}.
\end{equation}

\myparagraph{M\"{o}bius addition}Vector addition is not well-defined in the hyperbolic space and adding the vectors of two points straightly like Euclidean in the Poincare ball might result in a point outside the ball.
In this case, the M\"{o}bius addition~\cite{ganea2018hyperbolic} provides an analogue to the Euclidean addition in the hyperbolic space.
Here, $\oplus_{c} $ represents the M\"{o}bius addition as:
\begin{equation}
{\vh}_{i} \oplus_{c} \vh_{j} =\frac{\left(1+2 c\left\langle {\vh}_{i},  {\vh}_{j}\right\rangle+c\left\| {\vh}_{j}\right\|^{2}\right)  {\vh}_{i}+\left(1-c\left\| {\vh}_{i}\right\|^{2}\right)  {\vh}_{j}}{1+2 c\left\langle {\vh}_{i},  {\vh}_{j}\right\rangle+c^{2}\left\| {\vh}_{i}\right\|^{2}\left\|{\vh}_{j}\right\|^{2}}.
\end{equation}

\section{Methodology}
\begin{figure}[t]
	\centering
	\includegraphics[width=\linewidth]{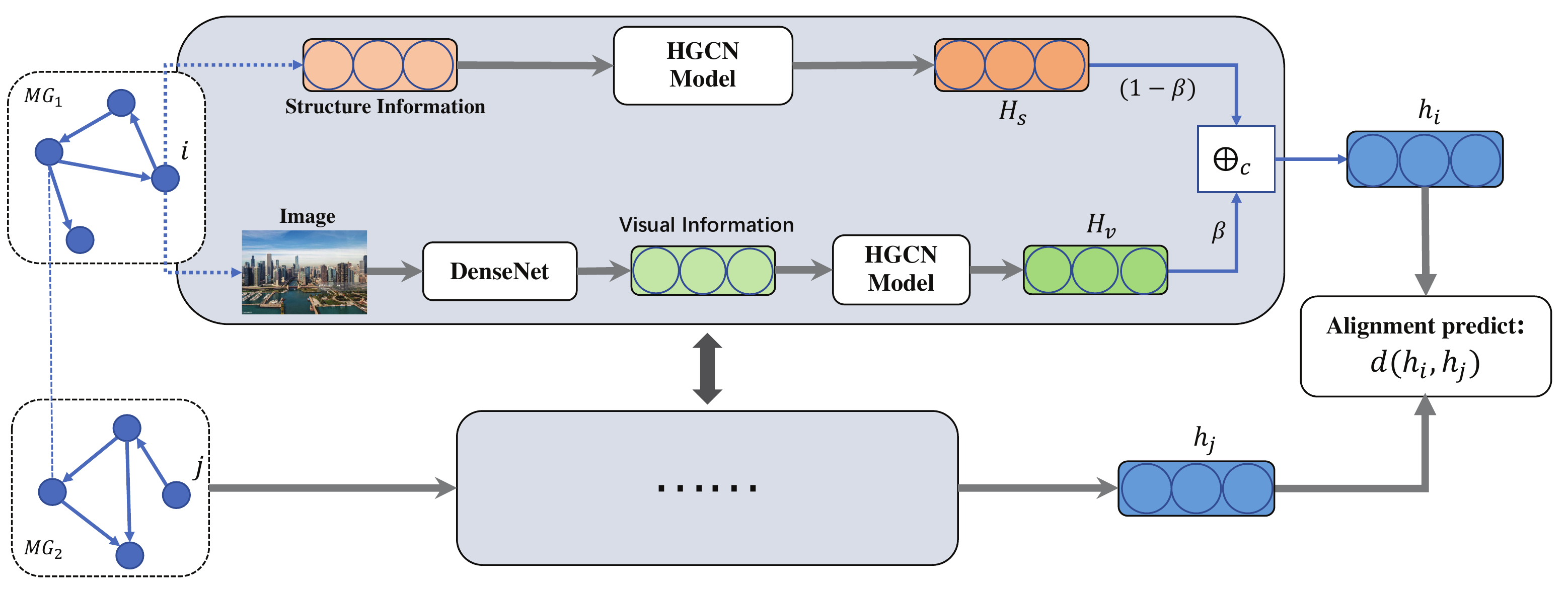}
	\caption{The framework of our proposed method.}
	\label{2}
\end{figure}

In this section, we present our proposed approach \our, which operates in the hyperbolic space. The framework is shown in Figure~\ref{2}. 
We first adopt \hgcn to learn the structural embeddings of entities. 
Then, we convert the images associated with entities into visual embeddings using the \densenet model, which are also projected into the hyperbolic space.
Finally, we combine these embeddings in the hyperbolic space and predict the alignment results using a pre-defined hyperbolic distance. 
We use Example~\ref{eg1} to illustrate our proposed model. 

\begin{Example}
\label{eg1}
Further to Example~\ref{eg}, by using structural information, it is easy to detect that \texttt{Nolan} in $MG_1$ is equivalent to \texttt{Nolan} in $MG_2$.
Nevertheless, merely using structural information is not enough, which might wrongly align \texttt{Michael Caine} in $MG_1$ to \texttt{Christan Bale} in $MG_2$.
In this case, the visual information would be very useful, since the images of \texttt{Michael Caine} in $MG_1$ and \texttt{Christan Bale} in $MG_2$ are very different. 
Therefore, we take into account both structural and visual information for alignment.
\end{Example}

In the following, we introduce the components of our proposal in detail.

\subsection{Structural Representation Learning}

We learn the structure representation of MMKGs by Hyperbolic Graph Convolutional Neural Networks, extending convolutional computation to manifold space, which benefits from the expressiveness of both graph neural networks and hyperbolic embeddings. More specifically, we first map input Euclidean features into hyperboloid manifold. Then, through \emph{feature transformation}, \emph{message passing} and \emph{non-linear activation} in the hyperbolic space, we can get the hyperbolic structural representations.


\myparagraph{Mapping input features to hyperboloid manifold}
In general, the input node features are produced by pre-trained Euclidean neural networks, and hence, they exist in the Euclidean space. 
To make the features available in the hyperbolic space, we first derive a mapping from Euclidean features to hyperbolic space.

Here, we assume that the input Euclidean features ${{\vx}^{E}} \in T_\mathbf{o}H_c$, where  $T_\mathbf{o}H_c$ represent the tangent space referring to $\mathbf{o}$, and $\mathbf{o} \in H_c$ denotes the north pole (origin) in hyperbolic space. 
We obtain the hyperbolic feature matrix {${\vx}^H$} via: 
$ {\vx}^H = \operatorname{exp}_o^c({\vx}^{E})$, where $\operatorname{exp}_o^c(\cdot)$ is defined in Equation~\ref{exp}.

\myparagraph{Feature transformation and propagation}
Similar to \gcn, feature transformation and message passing are also the core operations in hyperbolic structural learning. The operations are well-understood in the Euclidean space, however, their counterparts in hyperboloid manifold are non-trivial. 
In this connection, we could execute the functions with trainable parameters in the \emph{tangent space} of a point in the hyperboloid manifold, since the tangent space is Euclidean.
To this end, we leverage the $\operatorname{exp(\cdot)}$ map and $\operatorname{log(\cdot)}$ map to transform between hyperboloid manifold and the tangent space, so that we can use the tangent space $T_\mathbf{o}H_c^d$ to perform Euclidean operations.
 
We first use the logarithmic map to project hyperbolic representation ${\vx}_v^H \in R^{1 \times d}$ of node $v$ to the tangent space $T_\mathbf{o}H_c^d$. And in $T_\mathbf{o}H_c^d$, feature transformation and propagation rule for $v$ is calculated as:
\begin{equation}
{\vx}_v^T = \hat \mA \log _{\mathbf{o}}^{c}\left({\vx}_v^H\right)\mW,
\end{equation}
where $\vx_v^T \in R^{1\times d'} $ denotes the feature representation in the tangent space and $\hat \mA$ represents the symmetric normalized adjacency matrix; $\mW$ is a $ d' \times d $ trainable weight matrix.


\myparagraph{Non-linear activation with different curvatures}
After getting the representation in the tangent space, we use a non-linear activation function $\sigma^{\otimes^{c_{l}, c_{l+1}}}$ to learn non-linear transformations.
More concretely,  in $l$ layer's tangent space $ T_\mathbf{o} H^{d}_{c_{l}} $, we conduct Euclidean non-linear activation. Then we map it to the manifold of the next layer:
\begin{equation}
\sigma^{\otimes^{c_{l}, c_{l+1}}}\left(\vx_v^T\right)=\exp _{\mathbf{o}}^{c_{l+1}}\left(\sigma\left(\log _{\mathbf{o}}^{c_{l}}\left(\vx_v^T\right)\right)\right),
\end{equation}
where $-1/c_{l}$, $-1/c_{l+1}$ are hyperbolic curvatures at layer $l$ and $l+1$, respectively; the activation function $\sigma$ is chosen as $\operatorname{ReLU}(\cdot)$.
This step is significant as it allows us to smoothly vary the curvature at each layer, which is crucial to the overall performance due to the limited machine precision and normalization. 

Based on the hyperboloid feature transformation and non-linear activation, the convolutional computation in the hyperbolic space is redefined as:

\begin{equation}
 \begin{split}
 {\mH}^{l+1} = \exp _{\mathbf{o}}^{c_{l+1}}(\sigma(\hat \mA \log _{\mathbf{o}}^{c_l}\left({\mH}^{l}\right)\mW)),
 \end{split}
\end{equation}
where ${\mH}^{l+1}\in R^{n \times d^{l+1}}$, ${\mH}^{l} \in R^{n \times d^l}$ are the learned node embeddings in the hyperbolic space at $l+1$ layer and $l$ layer; and ${\mH}^{0} = {\vx}^{H} $; $\hat \mA$ represents the symmetric normalized adjacency matrix; $\mW$ is a $ d^l \times d^{l+1} $ trainable weight matrix.

\subsection{Visual Representation Learning}
We adopt the \densenet model~\cite{huang2017densely} to learn image embeddings, which is pre-trained on the ImageNet dataset~\cite{DBLP:conf/cvpr/DengDSLL009}. 
We remove the softmax layer in \densenet and obtain the 1920-dimensional embeddings for all images in the MMKGs.
Then, we project the embeddings into the hyperbolic space using \hgcn to improve their expressiveness.

\subsection{Multi-modal Information Fusion}
Both of the visual and structural information can contribute to the alignment results, as shown in Example~\ref{eg1}. Therefore, we design a novel method to combine \emph{structure information} and \emph{visual information} of MMKGs.
More specifically, we obtain the merged representation of entity $\mathbf{e}_i$ in hyperbolic space via:
\begin{equation}
{\vh}_{i} =(\beta \cdot{\mH}_{s}^i) \oplus_{c} ((1-\beta) \cdot {\mH}_{v}^i),
\end{equation}
where ${\mH}_{s}$ and ${\mH}_{v}$ are structure and visual embeddings learned from \hgcn model, respectively; $\beta$ is a hyper-parameter to balance the significance of these two features; the operator $\oplus_c$ is the M\"{o}bius addition. The combination requires the dimensions of structural and visual representations to be the same.

\subsection{Alignment Prediction}
We predict the alignment results based on the distance between entity representations from two MMKGs.
The Euclidean distance and Manhattan distance are commonly used distance measures in the Euclidean space~\cite{DBLP:conf/acl/WuLFWZ20,DBLP:conf/emnlp/LiCHSLC19}.
Nevertheless, in the hyperbolic space, we have to utilize the hyperbolic distance between nodes as the distance measure. 
For entities $e_i$ in $MG_1$ and $e_j$ in $MG_2$, the distance is defined as:
\begin{equation}
d_{c}\left({\vh}_{i}, {\vh}_{j}\right)= ||(-{\vh}_{i}) \oplus_{c} {\vh}_{j}||,
\end{equation}
where ${\vh}_{i}$ and ${\vh}_{j}$ denote the merged embeddings of $e_i$ and $e_j$ in the hyperbolic space, respectively; 
$\| \cdot \|$ is the $L_1$ norm;
the operator $\oplus_c$ is the M\"{o}bius addition.

The distance is expected to be small for equivalent entities and large for non-equivalent ones. 
For a specific entity $e_i$ in $MG_1$, our approach computes the distances between $e_i$ and all the entities in $MG_2$, and returns a list of ranked entities as candidate alignments.

\subsection{Model Training}
In order to embed equivalent entities as close as possible in the vector space, we use a set of known entity alignments (seed entities) $S$ as training data to train the model. Concretely, the model training is performed by minimizing the following margin-based ranking loss function: 
\begin{equation}
\begin{aligned} L =& \sum_{(e, v) \in S} \sum_{(e^{\prime}, v^{\prime}) \in S_{(e,v)}^{\prime}}\left[d_{c}\left({\vh}_{e}, {\vh}_{v}\right)+\gamma - d_{c}\left({\vh}_{e^{\prime}}, {\vh}_{v^{\prime}}\right)\right]_{+} \end{aligned},
\end{equation}
where $[x]_+ = \max\{{0,x}\}$; $(e,v)$ represents a seed entity pair and $S$ is the set of entity pairs;
$S_{(e,v)}^\prime$ denotes the set of negative instances constructed by corrupting $(e, v)$, i.e., replacing $e$ or $v$ with a randomly chosen entity in $MG1$ or $MG2$; $\gamma > 0$ denotes the margin hyper-parameter separating positive and negative instances. The margin-based loss function requires that the distance between the entities in positive pairs should be small, and the distance between the entities in negative pairs should be large.

\section{Experiment}

\subsection{Dataset and Evaluation Metric}
In the experiment, we use the datasets built in \cite{liu2019mmkg}, which are extracted from FreeBase, DBpedia, and YAGO respectively. These datasets use \texttt{FB15K} as a starting point to create the multi-modal knowledge graphs. Then they align the entities in \texttt{FB15K} with entities in others knowledge graphs using the reference links, resulting in \texttt{DB15K} and \texttt{YAGO15K}. We conducted experiments on two pairs of MMKGs, namely, \texttt{FB15K-DB15K} and \texttt{FB15K-YAGO15K}.

\begin{table}[H]
	\centering
	\caption{Statistic of the MMKGs Datasets.}
	\label{mdata}
	\begin{tabular}{c|c|c|c|c|c}
		\hline
		Datasets & Entities & Relations & Rel.Triples &Images & SameAs \\
		\hline
		\texttt{FB15K}  & 14,951&1,345&592,213&13,444& ~\\
		\hline
		\texttt{DB15K} &14,777&279&99,028&12,841&12,846\\
		\hline
		\texttt{YAGO15K} &15,404&32&122,886&11,194&11,199\\
		\hline
	\end{tabular}
	
\end{table}

Since the datasets do not provide the original pictures, to obtain the relevant images for each entity, we use the URIs built in~\cite{moon2018multimodal} and implement a web crawler that is able to parse query results from the image search engines, i.e., Google Images~\myfootnote{https://www.google.com/imghp?hl=EN}, Bing Images~\myfootnote{https://www.bing.com/image}, and Yahoo Image Search~\myfootnote{https://images.search.yahoo.com/}. 
Then, we assign the pictures obtained by different search engines to different MMKGs, so as to reflect the heterogeneity of different MMKGs. 

Table~\ref{mdata} outlines the detailed information of the datasets. Each dataset contains nearly 15 thousand entities and more than 11 thousand image sets of entity.
The \emph{Images} column represents the number of entities that possess the image sets.
These alignments are given by the $SameAs$ predicates that have been previously found. In the experiments, the known equivalent entity pairs are used for model training and testing.


\myparagraph{Evaluation metric} We use $Hits@k$ as the evaluation measure to assess the performance of all the approaches. $Hits@k$ measures the proportion of correct aligned entities ranked in the top-$k$ candidates.

\subsection{Experiment Setting and Competing Approaches}

\myparagraph{Experiment Setting} 
In order to assess the performance under different percentages of the given alignments $P($\%$)$, we evaluate the methods with low ($20\%$), medium ($50\%$) and high percentage ($80\%$) of the given seed entity pairs. The remaining $sameAs$ triples are used for test. 
For fairness, we keep the number of dimensions to be the same (i.e., 400) for \gcnalign and \our. The other parameters of \gcnalign follow~\cite{wang2018cross}.
For the parameters of our approach \our, we generate 6 negative samples for each positive one; the margin hyper-parameters in the loss function are $\gamma_{\our-s} = 0.5$ and  $\gamma_{\our-v} = 1.5$, respectively. We optimize \our with $Adam$.

\myparagraph{Competing approaches} To demonstrate the superiority of our proposed model, we chose three state-of-the-art approaches as competitors:
\begin{itemize}
	\item \gcnalign~\cite{wang2018cross} adopts \gcn to encode the structural information of entities, and then combines relation and image embeddings for the entity alignment task.
	
	\item \poe~\cite{liu2019mmkg} is the product of experts model. It computes the scores of facts under each modality, and learns the entity embeddings for entity alignment. \poe combines information from two modalities. We also compare with the variant \poe-s, which merely uses the structure information.
	\item \ikrl~\cite{xie2017image-embodied} integrates image representations into an aggregated image-based representation via an attention-based method. It is originally proposed in the field of knowledge representation and we adopt it to tackle MMEA.
\end{itemize}

To further demonstrate the benefit from hyperbolic geometry, especially in learning structural features, we conduct exploratory experiments by solely using \emph{structure information} for EA, resulting in \our-s, \gcnalign-s and \poe-s. 
In addition, to evaluate the contribution of \emph{visual information}, we compare \poe, \gcnalign and \our with just \emph{visual information}, namely \poe-v, \gcnalign-v and \our-v.

\subsection{Results}

The results are shown in Table~\ref{mresult}. It is obvious that \our achieves the best performance in all cases. Especially in \texttt{FB15K-YAGO15K}, with 80\% seed entity pairs, \our outperforms \poe and \gcnalign by nearly 15\% in terms of $Hits@1$. With 20\% seed entity pairs, our approach also shows better results and the improvement of $Hits@1$ is around 2\% and $Hits@10$ is up to 20\%.
According to the results of \poe, we can find that there is little enhancement from $Hits@1$ to $Hits@10$, ranging from 4\% to 9\%. In contrast, the enhancements from $Hits@1$ to $Hits@10$ of \our are at least 20\% in all situations. By the way, \our outperforms \ikrl by a large margin.

\begin{table}[H]
	\centering
	\caption{Alignment prediction on both datasets for different percentages of P. }
	\label{mresult}
	\resizebox{0.9\textwidth}{!}{
		\begin{tabular}{c|cc|cc|cc}
			\hline
			{\multirow{2}*{\texttt{FB15K-DB15K}}} & \multicolumn{2}{c|}{20\%} & \multicolumn{2}{c|}{50\%} & \multicolumn{2}{c}{80\%} \\
			~ & $Hits@1$ & $Hits@10$ & $Hits@1$ & $Hits@10$ & $Hits@1$ & $Hits@10$ \\
			\hline
			\poe & 11.1 & 17.8& 23.5 & 33.0 & 34.4 & 40.6 \\
			\hline
			\gcnalign & 5.35 &17.11 &13.85 &34.31&22.18 & 48.95 \\
			\hline
			\ikrl &1.01 & 2.40 & 2.77 & 5.79 &5.41 &11.09\\
			\hline
			\our &\textbf{12.65} &\textbf{36.86}& \textbf{26.23}& \textbf{58.08}& \textbf{41.68}&\textbf{78.55} \\
			\hline
			\hline
			{\multirow{2}*{\texttt{FB15K-YAGO15K}}} & \multicolumn{2}{c|}{20\%} & \multicolumn{2}{c|}{50\%} & \multicolumn{2}{c}{80\%} \\
			~ & $Hits@1$ & $Hits@10$ & $Hits@1$ & $Hits@10$ & $Hits@1$ & $Hits@10$ \\
			\hline
			\poe & 8.7 & 13.3& 18.5 & 24.7 & 28.9 & 34.3 \\
			\hline
			\gcnalign & 6.76 &17.99 &16.47 &35.85& 28.75&53.05 \\
			\hline
			\ikrl & 0.86 & 1.75 &1.95 &3.73 &3.57 &7.14 \\
			\hline
			\our &\textbf{10.51} &\textbf{31.27}& \textbf{26.50}& \textbf{58.08}& \textbf{43.30}&\textbf{80.11} \\
			\hline
	\end{tabular}}

\end{table}

As shown in Table~\ref{mresult2}, if solely using \emph{structural information}, \our-s still leads to better results than other two methods. More concretely, our proposed approach outperforms \gcnalign-s by nearly 5\% on \texttt{FB15K-DB15K} and 3\% on \texttt{FB15K-YAGO15K} in terms of $Hits@1$ with 20\% seed alignments. 
With 50\% and 80\% seed entity pairs, \our-s brings significantly better results. The improvements range from 10\% to 18\% regarding $Hits@1$ and from 20\% to 30\% in terms of $Hits@10$. According to the results, it can be concluded that our approach does have advantage in capturing accurate hierarchical structure representation.

\begin{table}[H]
	\centering
	\caption{Results of three methods with \emph{structure information}. }
	\label{mresult2}
	\resizebox{0.9\textwidth}{!}{
		\begin{tabular}{c|cc|cc|cc}
			\hline
			{\multirow{2}*{\texttt{FB15K-DB15K}}} & \multicolumn{2}{c|}{20\%} & \multicolumn{2}{c|}{50\%} & \multicolumn{2}{c}{80\%} \\
			~ & $Hits@1$ & $Hits@10$ & $Hits@1$ & $Hits@10$ & $Hits@1$ & $Hits@10$ \\
			\hline
			\poe-s & 10.7 & 16.5& 22.9 & 31.7 & 33.6 & 38.6 \\
			\hline
			\gcnalign-s & 5.35 &17.11 &13.85 &34.31&22.18 & 48.95 \\
			\hline
			\our-s &  \textbf{11.73} &  \textbf{33.56} & \textbf{24.84}& \textbf{56.69}& \textbf{40.87}& \textbf{76.77} \\
			\hline
			\hline
			{\multirow{2}*{\texttt{FB15K-YAGO15K}}} & \multicolumn{2}{c|}{20\%} & \multicolumn{2}{c|}{50\%} & \multicolumn{2}{c}{80\%} \\
			~ & $Hits@1$ & $Hits@10$ & $Hits@1$ & $Hits@10$ & $Hits@1$ & $Hits@10$ \\
			\hline
			\poe-s & 8.4 & 12.3& 18.0 & 23.1 & 28.1 & 31.9 \\
			\hline
			\gcnalign-s & 6.76 &17.99 &16.47 &35.85& 28.75&53.05 \\
			\hline
			\our-s & \textbf{9.66} & \textbf{28.96} & \textbf{25.37}& \textbf{56.60}& \textbf{42.63}& \textbf{78.42} \\
			\hline
		\end{tabular}
	}
\end{table}

With \emph{visual information} (Table~\ref{mresult3}), we compare three variants: \poe-v, \gcnalign-v and \our-v. It is obvious that \gcnalign does not obtain useful visual representation for MMEA. If solely using structural information, \our-v still shows better results than \poe-v. More concretely, our proposed approach outperforms \poe-v slightly with 20\% seed alignments, less than 1\% on both datasets regarding $Hits@1$.
With 80\% seeds on \texttt{FB15K-DB15K}, \our-v brings significantly better results. The improvements are around 7\% regarding $Hits@1$ and 18\% in terms of $Hits@10$. The results demonstrate the effectiveness of our proposed mode for learning visual features.
\begin{table}[H]
	\centering
	\caption{Comparison of three methods with \emph{visual information}.}
	\label{mresult3}
	\resizebox{0.9\textwidth}{!}{
	\begin{tabular}{c |c c|c c|c c}
		\hline
		{\multirow{2}*{\texttt{FB15K-DB15K}}} & \multicolumn{2}{c|}{$20\%$} & \multicolumn{2}{c|}{50\%} & \multicolumn{2}{c}{80\%} \\
	     ~ & $Hits@1$ & $Hits@10$ & $Hits@1$ & $Hits@10$ & $Hits@1$ & $Hits@10$ \\
		\hline
		\poe-v & 0.8& 2.7& 1.3&3.8&1.7&5.9 \\
		\hline
		\gcnalign-v & 0.0& 0.0 &0.0 &0.0  &0.0 & 0.0 \\
		\hline
		\our-v &\textbf{1.77} &\textbf{8.08}& \textbf{3.33}& \textbf{12.65}& \textbf{9.05}&\textbf{24.20} \\
		\hline
		\hline
		{\multirow{2}*{\texttt{FB15K-YAGO15K}}} & \multicolumn{2}{c|}{$20\%$} & \multicolumn{2}{c|}{50\%} & \multicolumn{2}{c}{80\%} \\
		 ~ & $Hits@1$ & $Hits@10$ & $Hits@1$ & $Hits@10$ & $Hits@1$ & $Hits@10$ \\
		\hline
	    \poe-v & 0.7& 2.4& 1.1&3.2&1.7&5.5 \\
		\hline
		\gcnalign-v & 0.0& 0.0 &0.0 &0.0  &0.0 & 0.0 \\
		\hline
		\our-v &\textbf{1.35} &\textbf{5.43}& \textbf{2.71}& \textbf{11.15}& \textbf{5.79}&\textbf{18.07} \\
		\hline
	\end{tabular}
	}
\end{table}

\subsection{Ablation Experiment}
MMKGs consist of information in multiple modalities. We take the structural and visual information into account in this work.
In order to further validate the effectiveness of multi-modal knowledge for MMEA, we conduct the ablation experiment. 
By comparing \our and \our-s in Table~\ref{mresult} and Table~\ref{mresult2}, it can be seen that adding \emph{visual information} in our approach does lead to slightly better results and the improvements are around 1\% in terms of $Hits@1$.
Moreover, by comparing \our and \our-v in Table~\ref{mresult} and Table~\ref{mresult3}, we can also see that the structural information is of great significance. 
The ablation study shows that MMEA mainly relies on the \emph{structural information}, but the \emph{visual information} is still useful. 
In addition, it demonstrates that the combination of these information works much better.



{\subsection{Case Study}}

One key property of hyperbolic spaces is that they expand faster than Euclidean spaces, because Euclidean spaces expand polynomially while hyperbolic spaces expand exponentially. In other words, for the neighbor nodes of the central node, they are distributed in bigger space and the distances between them are farther, which can help distinguish similar entities.

To further demonstrate the effectiveness of the embeddings in hyperbolic space, we conduct the following case study. We choose \texttt{Michael Caine} as the root node. We visualize the embeddings of its 1-hop film-related entities learned from \gcnalign and \our separately in the PCA-projected spaces in Figure 4. It can be observed that, for the entities of the same type or with similar structure information, especially for entity \texttt{Alfie} and \texttt{B-o-B}, their Euclidean embeddings (generated via \gcnalign) are placed closely. While in Hyperbolic space, the distances between such entities are relatively farther away (with only a few exceptions). This validates that the hyperbolic structure representation can help distinguish similar entities. Moreover, by placing similar entities (in the same KG) distantly, the hyperbolic representation can help the alignment process (alignment across KG).



{
 For instance, as shown in Figure 4(a), for entity \texttt{Alfie} in \texttt{FB15K}, the closest entity to it is entity \texttt{B-o-B} (which is incorrect). However, in Figure 4(b), for entity \texttt{Alfie}, the entity \texttt{B-o-B} is placed far away from it, and the closest entity to it is \texttt{Alfie} in \texttt{DB15K}. Hence, by using the hyperbolic projection, similar entities in the same KG are well distinguished and placed far away, such that alignment mistakes could be avoided.
}

\begin{figure}[H]
\centering

 \subfigure[Embedding generated from \gcnalign]{\includegraphics[width=0.78\hsize]{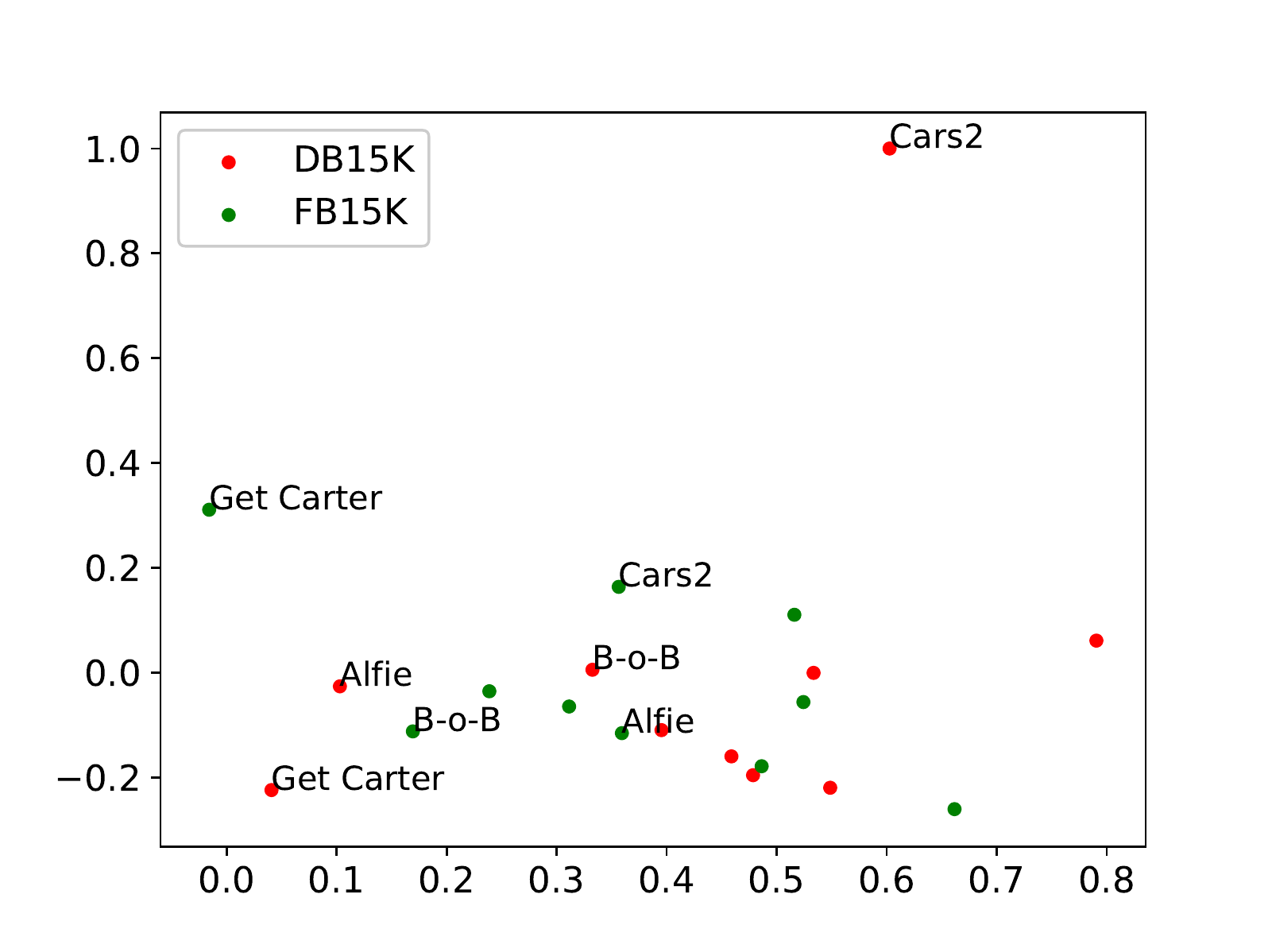}}  
 \subfigure[Embedding generated from \our]{\includegraphics[width=0.78\hsize]{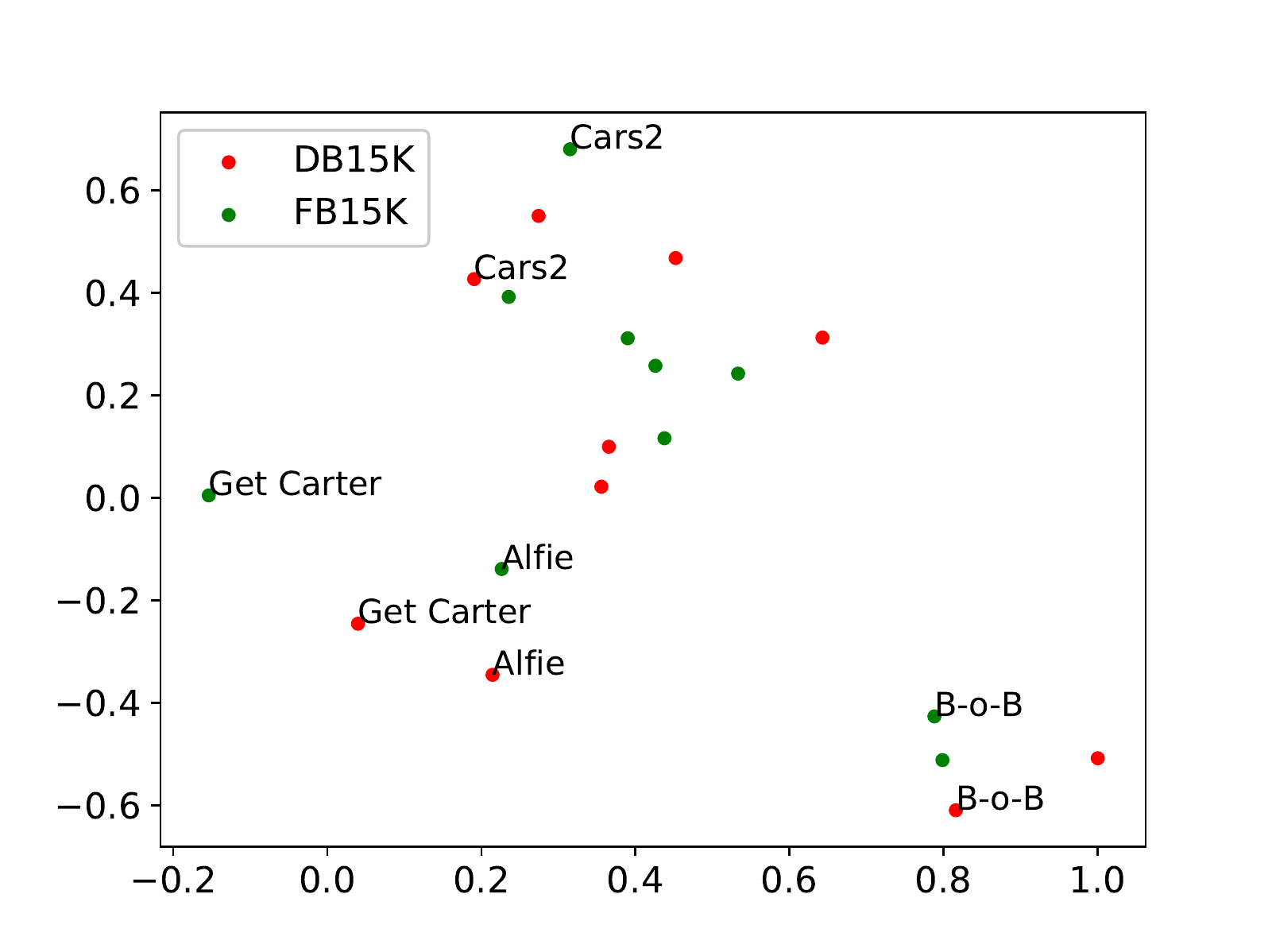}}

\centering

\caption{
The embeddings of 1-hop film-related neighbor entities of \texttt{Michael Caine} generated from \gcnalign and \our separately in the PCA-projected space.
The green points represent entities in FB15K; red points represent entities in DB15K. For simplicity, we annotate part of entities. \texttt{B-o-B} is the abbreviation of \texttt{Battle of Britain}.
}

\end{figure}



\subsection{Additional Experiment}

\begin{table}[H]\footnotesize
	\centering
	\caption{\textbf{Details of the cross-lingual datasets}}
	\label{cdata}
	\resizebox{0.9\textwidth}{!}{
		\begin{tabular}{c|c|c c c c c} 
			\hline
			\multicolumn{2}{c|}{Datasets} &Entities & Relations &Attributes &Rel.triples &Attr.triples \\
			\hline
			\multirow{2}*{\dbpsz} & Chinese &66,469 &2,830 &8,113&153,929&379,684\\
			~ & English & 98,125 &2,137 &7,173&237,674&567,755\\
			\hline
			\multirow{2}*{\dbpsj} & Japanese &65,744 &2,043 &5,882&164,373&354,619\\
			~ & English & 95,680 &2,096 &6,066&233,319&497,230\\
			\hline
			\multirow{2}*{\dbpsf} & French &66,858 &1,379 &4,547&192,191&528,665\\
			~ & English & 105,889 &2,209 &6,422&278,590&576,543\\
			\hline
		\end{tabular}
	}
\end{table}
The cross-lingual EA datasets are the most widely-used datasets to assess EA approaches.  We added the experiments on them to show that our proposed approach can work on such popular datasets (and the cross-lingual EA task). Note that diverse languages are not taken as multiple modalities, and the cross-lingual EA is in essence single-modal EA.
We use the \texttt{DBP15K} datasets in the experiments, which were built by \cite{sun2017cross}. 
The datasets were generated from DBpedia, containing rich inter-language links between different language versions. 
Each dataset contains data in different languages and 15 thousand known inter-language links connecting equivalent entities in two KGs, which are used for model training and testing. 
Following the setting in~\cite{wang2018cross}, we use $30\%$ of inter-language links for training, and $70\%$ of them for testing. $Hits@k$ is used as the evaluation measure.

Both the dimensions of structure and attribute are set to 300-dimension for \gcnalign. 
\gcnalign-s and \our-s represent adopting structure information;
\gcnalign-a and \our-a represent adopting attribute information; and \gcnalign and \our combine both the structure information and attribute information. 

As shown in Table~\ref{tab:cross}, in all datasets, \our-s outperforms \gcnalign-s and the improvements are around 7\% in terms of $Hits@1$ and more than 10\% in terms of $Hits@10$. It demonstrates that \our benefits from hyperbolic geometry and is able to capture better structural features.
In addition, with the combination of structure information and attribute information, our proposed approach outperforms \gcnalign by around 10\% regarding $Hits@1$.
As for the attribute information, \our-a also gets significantly better results than \gcnalign-a, which increases by around 15\% on $Hits@1$ in all datasets.

\begin{table}[H]
	\footnotesize
	\centering
	\caption{Result in Cross-lingual datasets.}
	\label{tab:cross}
	\begin{tabular}{c|c c|c c}
		\hline
		\multirow{2}*{\dbpsz} &\multicolumn{2}{c|}{\texttt{ZH-EN}} & \multicolumn{2}{c}{\texttt{EN-ZH}}\\
		~  & $Hits@1$ & $Hits@10$ & $Hits@1$ & $Hits@10$\\
		\hline
		
		\gcnalign-s &39.42 &71.34 &33.60 &65.23 \\
		\our-s &\textbf{46.23} &\textbf{82.36} &\textbf{44.53} &\textbf{81.95} \\
		\hline
		\gcnalign-a &13.44 &40.94 &12.54 &38.78 \\
		\our-a&\textbf{33.99} &\textbf{71.15} &\textbf{32.80} &\textbf{69.79} \\
		\hline
		\gcnalign &43.08 &75.92 &36.25 &69.17 \\
		\our&\textbf{54.04} &\textbf{87.88} &\textbf{51.88} &\textbf{86.57} \\
		\hline
		\hline
		\multirow{2}*{\dbpsj} &\multicolumn{2}{c|}{\texttt{JA-EN}} & \multicolumn{2}{c}{\texttt{EN-JA}}\\
		
		~  & $Hits@1$ & $Hits@10$ & $Hits@1$ & $Hits@10$\\
		\hline
		
		\gcnalign-s &39.95 &72.72 &36.09 &67.43 \\
		\our-s & \textbf{47.63}& \textbf{83.96}&\textbf{47.24} &\textbf{83.96} \\
		\hline
		\gcnalign-a &9.27 &31.85 &8.78 &31.89 \\
		\our-a&\textbf{28.36} &\textbf{63.99} &\textbf{27.73} &\textbf{63.97} \\
		\hline
		\gcnalign & 42.51&75.74 &38.31 &70.49 \\
		\our& \textbf{53.06}&\textbf{87.47} &\textbf{52.65} &\textbf{87.41} \\
		\hline
		\hline
		{\multirow{2}*{\dbpsf}} &\multicolumn{2}{c|}{\texttt{FR-EN}} & \multicolumn{2}{c}{\texttt{EN-FR}}\\
		
		~  & $Hits@1$ & $Hits@10$ & $Hits@1$ & $Hits@10$\\
		\hline
		\gcnalign-s &38.38 &74.45 &37.37 &71.65 \\
		\our-s &\textbf{44.27} &\textbf{83.15} &\textbf{43.81} &\textbf{83.14}  \\
		\hline
		\gcnalign-a & 2.65&13.50 &3.02 &14.51 \\
		\our-a& \textbf{12.40}&\textbf{48.70} &\textbf{15.44} &\textbf{52.12} \\
		\hline
		\gcnalign & 39.48&76.05 &38.44 &73.33 \\
		\our& \textbf{48.40}&\textbf{86.49} &\textbf{48.15} &\textbf{86.18} \\
		\hline
		
	\end{tabular}
\end{table}

\section{Conclusion}
In this paper,  we propose \our, a novel multi-modal EA approach which effectively integrates multi-modal information for EA in MMKGs.
It extends the Euclidean representation to hyperboloid manifold and adopts \hgcn to learn structural representations of entities. A more advanced model \densenet is leveraged to learn better visual representations. The structural and visual embeddings are further aggregated in the hyperbolic space to predict potential alignments.
We validate the effectiveness of our proposed model via comprehensive experimental evaluations. Additional experiments also confirm that \hgcn obtains better structural features of knowledge graphs in the hyperbolic space.








\bibliographystyle{abbrv}

\end{document}